%% file: main.tex
\documentclass[10pt,twocolumn,letterpaper]{article}

\usepackage{cvpr}              
\usepackage{threeparttable}
\input{preamble}

\definecolor{cvprblue}{rgb}{0.21,0.49,0.74}
\usepackage[pagebackref,breaklinks,colorlinks,allcolors=cvprblue]{hyperref}

\def\confName{CVPR}
\def\confYear{2026}

\usepackage{amsmath} 
\usepackage{amssymb}  
\usepackage{booktabs}
\usepackage{multirow}
\usepackage{kotex}
\usepackage{lipsum}
\usepackage{float}
\usepackage{amsmath}
\usepackage{colortbl}
\usepackage{threeparttable}
\usepackage{graphicx}    
\usepackage{caption} 
\usepackage{booktabs}
\usepackage[capitalize]{cleveref}
\usepackage{placeins}
\usepackage{xcolor}
\definecolor{mybrown}{rgb}{0.65, 0.16, 0.16} 

\title{Rethinking Pose Refinement in 3D Gaussian Splatting under Pose Prior and Geometric Uncertainty}

\author{
    Mangyu Kong$^{1}$\quad
    Jaewon Lee$^{1}$\quad
    Seongwon Lee$^{2^\star}$\quad
    Euntai Kim$^{1,3\thanks{Corresponding authors.}}$\\[1.5mm] 
    $^{1}$Yonsei University\quad 
    $^{2}$Kookmin University\quad
    $^{3}$Korea Institution of Science and Technology
}

\begin{document}
\maketitle
\input{sec/0_abstract}

\input{sec/1_intro}
\input{sec/2_related}
\input{sec/3_method}

\input{sec/4_exps}

\input{sec/5_conclusion}
\input{sec/6_acknowledgement}

{
    \small
    \bibliographystyle{ieeenat_fullname}
    \bibliography{main}
}

\setcounter{section}{0} 
\setcounter{figure}{0}  
\setcounter{table}{0}   

\input{sec/X_suppl}



\end{document}

%% file: preamble.tex
\usepackage[accsupp]{axessibility}  

\usepackage{pifont}
\usepackage{multirow}
\usepackage{algorithm}
\usepackage{algpseudocode}

%% file: sec/0_abstract.tex
\begin{abstract}

3D Gaussian Splatting (3DGS) has recently emerged as a powerful scene representation and is increasingly used for visual localization and pose refinement. However, despite its high-quality differentiable rendering, the robustness of 3DGS-based pose refinement remains highly sensitive to both the initial camera pose and the reconstructed geometry. In this work, we take a closer look at these limitations and identify two major sources of uncertainty: (i) pose prior uncertainty, which often arises from regression or retrieval models that output a single deterministic estimate, and (ii) geometric uncertainty, caused by imperfections in the 3DGS reconstruction that propagate errors into PnP solvers. Such uncertainties can distort reprojection geometry and destabilize optimization, even when the rendered appearance still looks plausible.
To address these uncertainties, we introduce a relocalization framework that combines Monte Carlo pose sampling with Fisher Information–based PnP optimization. Our method explicitly accounts for both pose and geometric uncertainty and requires no retraining or additional supervision. Across diverse indoor and outdoor benchmarks, our approach consistently improves localization accuracy and significantly increases stability under pose and depth noise. Our code is available on the \href{https://kmk97.github.io/UGSLoc/}{project website}.

\end{abstract}

%% file: sec/1_intro.tex
\section{Introduction}
\label{sec:intro}

\begin{figure}[t]
  \centering
  \includegraphics[width=\columnwidth]{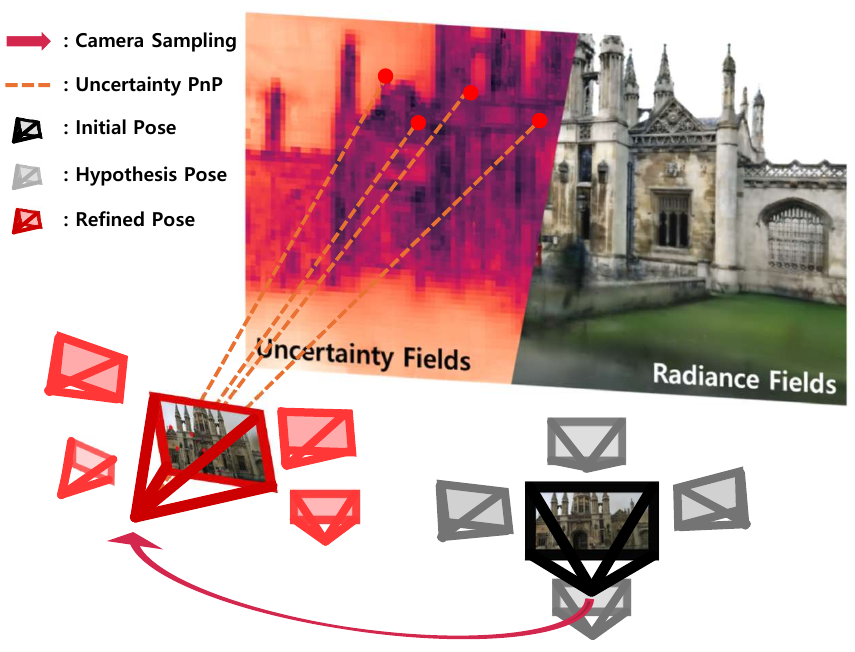}
\caption{\textbf{UGS-Loc.} Our proposed framework refines camera pose by incorporating pose prior uncertainty via Monte Carlo sampling and geometric uncertainty via Uncertainty fields, achieving robust localization without retraining.}
\label{fig:fig1}
\vspace{-0.5cm}
\end{figure}

\begin{figure*}[t!]
  \centering
  \includegraphics[width=\textwidth]{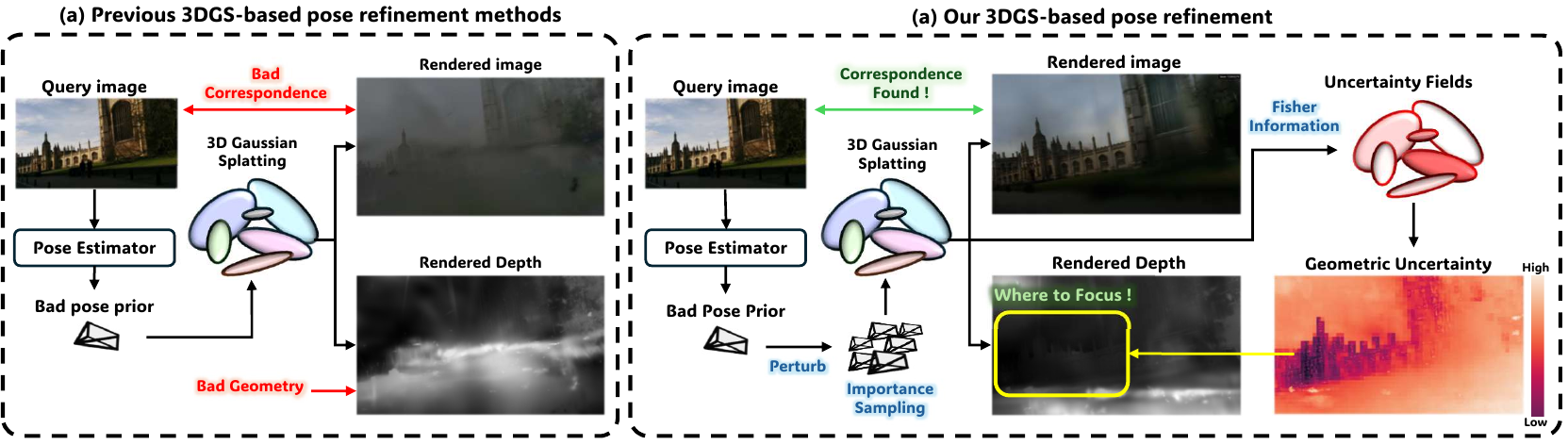}
\caption{\textbf{Comparison between Previous 3DGS-based Pose Refinement and Our Proposed UGS-Loc.} (a) Existing pipelines refine a single deterministic pose prior using rendered depth, which makes them vulnerable to inaccurate priors and geometric errors—leading to bad correspondences and unstable refinement. (b) Our method instead applies Monte Carlo pose sampling and geometric uncertainty from Fisher Information to guide correspondences and focus refinement on reliable regions, achieving robust pose estimation even under poor pose priors and uncertain geometry.}
\label{fig:compare}

\vspace{-0.5cm}
\end{figure*}

Visual localization aims to estimate the 6-DoF camera pose of a query image within a known scene and plays a key role in AR/VR, autonomous driving, and robotics. A main factor governing the performance of localization methods is how the scene is represented. Traditional structure-based pipelines rely on sparse 3D point clouds reconstructed via Structure-from-Motion~\cite{sarlin2019coarse,sattler2018benchmarking}. Local features extracted from the query image are matched to reference images or 3D points~\cite{sarlin2020superglue,detone2018superpoint,mur2015orb}, and the resulting correspondences are used to estimate the camera pose through PnP algorithms~\cite{gao2003complete,lepetit2009ep,fischler1981random}. Scene Coordinate Regression (SCR) approaches~\cite{brachmann2017dsac,brachmann2021visual,brachmann2023accelerated} extend this paradigm by predicting dense scene coordinates, and Absolute Pose Regression (APR) methods~\cite{kendall2015posenet,chen2022dfnet,li2024unleashing,chen2024map,shavit2021learning} learn to map images directly to poses. These approaches illustrate how advances in map representation, from sparse geometric maps to implicitly learned coordinate or pose models, continue to shape the design of localization systems.

Recently, neural scene representations such as NeRF~\cite{mildenhall2021nerf} and 3DGS~\cite{kerbl20233d} emerged as a new type of scene representation, and they triggered the development of new pose-refinement methods. These methods can be summarized as render-and-compare. They first render a synthetic view from an initial pose obtained by retrieval~\cite{arandjelovic2016netvlad,lee2022correlation}, APR, or SCR, and then align this view to the query image using photometric loss~\cite{yen2021inerf,sun2023icomma,lin2023parallel} or feature distances~\cite{trivigno2024unreasonable,chen2024neural,liu2024hr}. More recent methods~\cite{liu2024gs,huang2025sparse,li2024unleashing} extend this idea: they find dense 2D–2D correspondences~\cite{leroy2024grounding,sun2021loftr} between the query and the rendered view, lift these matches into 3D using rendered depth, and solve the pose with PnP-RANSAC. This correspondence-driven version of render-and-compare now leads pose refinement and achieves state-of-the-art relocalization accuracy in rendering-based pose refinement. However, when the pose refinement methods are applied to 3DGS, we believe that the existing pose refinement methods overlook some points and may degrade some performance.

3D Gaussian Splatting models geometry with explicit elliptical primitives, but the primitives are just approximate and the depth they render is not uniformly reliable. Regions reconstructed from sparse training views~\cite{xiong2023sparsegs} or contaminated by dynamic objects~\cite{kulhanek2024wildgaussians} often exhibit noticeable artifacts, leading to spatial variations in geometric accuracy. Despite these spatial variations, current refinement methods treat all rendered depths as equally trustworthy when lifting 2D–2D correspondences into 3D. As a result, they ignore the fact that reconstruction fidelity can vary significantly across different parts of the scene.

This challenge becomes even more pronounced when considering the dependence on an accurate pose prior. Since 3DGS-based refinement~\cite{liu2024gs,huang2025sparse,li2024unleashing} operates by rendering synthetic views from a single initial estimate, the quality of correspondence hinges on how well this pose is aligned with the true scene content. When the pose prior is biased or occluded by artifacts, the rendered viewpoints may exhibit limited visibility, causing feature matchers~\cite{leroy2024grounding,detone2018superpoint,sun2021loftr} to struggle to establish reliable correspondences.

To address these challenges, we propose UGS-Loc, an uncertainty-aware pose refinement framework tailored to 3D Gaussian Splatting. UGS-Loc represents the pose prior as a distribution and explores it through Monte Carlo sampling, enabling the refinement pipeline to recover even when the initial pose lies far from reliable regions of the scene. In parallel, we quantify geometric uncertainty directly from the Gaussian parameters via Fisher Information~\cite{fisherrf} and integrate this signal into the PnP-RANSAC process, guiding correspondence sampling toward geometrically trustworthy regions. By jointly modeling pose and geometric uncertainties inherent to 3DGS, UGS-Loc achieves robust, training-free localization that remains stable and consistent across challenging scene conditions.

%% file: sec/2_related.tex
\section{Related Works}
\label{sec:retlatingwork}
\paragraph{Visual Localization.} The task of visual localization is to determine the pose of a query image in relation to a reconstructed 3D scene. A commonly used pre-built representation of the scene is a 3D point cloud reconstructed via Structure-from-Motion (SfM)~\cite{zheng2015structure}.
Each 3D point in this cloud is associated with a set of visual descriptors from database images, allowing the system to establish reliable 2D–3D correspondences during localization. Query image keypoints are matched between the query image and the 3D model to establish 2D–3D correspondences~\cite{sarlin2019coarse,taira2018inloc,detone2018superpoint,sarlin2020superglue}. Scene coordinate regression (SCR) methods~\cite{wang2024glace,brachmann2023accelerated,brachmann2017dsac} train the network to acquire 2D-3D correspondences directly. The correspondences obtained from both methods are then fed into a PnP solver to estimate the camera pose. Alternatively, absolute pose regression (APR)~\cite{kendall2015posenet,shavit2021learning} predicts the 6-DoF camera pose directly from a query image through a neural network. However, these approaches often suffer from limited pose accuracy. Our method refines the poses estimated by SCR and APR methods using 3D Gaussian Splatting (3DGS) without requiring any additional training.

\paragraph{Radiance Field-based Visual Localization.} 
To overcome the sparsity and limited expressiveness of traditional 3D point cloud representations, recent works have explored dense scene representations such as radiance fields for visual localization.
Leveraging the powerful novel-view synthesis capability of Neural Radiance Fields (NeRF)~\cite{mildenhall2021nerf} and Gaussian Splatting (3DGS)~\cite{kerbl20233d}, these methods utilize photo-realistic rendering for both pose refinement and data augmentation.
For instance, radiance field–based augmentation enriches APR frameworks~\cite{chen2022dfnet,li2024unleashing,moreau2022lens} by generating synthetic training views across diverse viewpoints and lighting conditions, improving robustness and generalization to unseen domains.

A dominant line of research adopts the render-and-compare strategy, where the initial camera pose is iteratively refined by minimizing discrepancies between the query image and novel views rendered from the estimated pose.
iNeRF~\cite{yen2021inerf} pioneered this paradigm by optimizing the camera pose directly via inverse rendering in NeRF, inspiring a following works~\cite{lin2023parallel,sun2023icomma}.
These approaches optimizes poses directly using photometric~\cite{botashev2024gsloc} or feature-space alignment~\cite{chen2024neural} to improve localization precision. In parallel, correspondence-based pipelines have emerged as a strong alternative. CROSSFIRE~\cite{moreau2023crossfire} integrates specialized feature descriptors tailored to neural scene representations, enabling accurate 2D–3D correspondence establishment followed by PnP-RANSAC for pose estimation. This design~\cite{zhao2024pnerfloc,sidorov2024gsplatloc,zhou2024nerfect,liu2023nerf} significantly boosts localization accuracy by leveraging features that are deeply aligned with the underlying radiance field.

Beyond learning scene-specific descriptors, GS-CPR~\cite{liu2024gs} adopts a simpler and more efficient strategy: using a pose prior from APR or SCR, it renders high-quality synthetic views, performs direct 2D–2D matching with the query, lifts correspondences into 3D via rendered depth, and refines the pose through PnP optimization. Similar 2D-2D correspondence-driven refinement pipelines are also employed in several recent 3DGS-based localization methods~\cite{huang2025sparse,li2024unleashing}, where a coarse pose predicted by each method serves as the initial pose. Its simplicity and efficiency make this correspondence-based refinement pipeline a widely used baseline for 3DGS-based pose refinement. However, these refinement-based approaches still heavily rely on accurate pose priors or well-reconstructed scene geometry, making them vulnerable to uncertainty and geometric inconsistencies in practical scenarios.
In this work, we address these limitations by introducing a localization framework that handles uncertainty in both pose prior and scene geometry.

\paragraph{Uncertainty Quantification in Radiance Fields.}
Recent studies~\cite{yan2023active,yan2024cf,goli2024bayes,zhan2022activermap,fisherrf} have proposed diverse strategies to quantify uncertainty within Radiance Fields.
NeRF-based methods~\cite{zhan2022activermap,yan2023active,lee2022uncertainty} estimate the reliability of scene reconstruction by modeling uncertainty along casted rays, typically through ray-wise density distributions. Few-view uncertainty estimation methods~\cite{yan2024cf,shen2021stochastic} model epistemic uncertainty using variational inference. Bayes' Rays~\cite{goli2024bayes} estimates spatial uncertainty by applying a perturbation field over a spatial grid and interpolating uncertainty at query points. Similarly, FisherRF~\cite{fisherrf} computes Fisher information over NeRF and 3DGS to measure view-dependent uncertainty.

However, these uncertainty‐quantification approaches are primarily designed for improving scene reconstruction, either by filtering out outliers~\cite{goli2024bayes,kulhanek2024wildgaussians,martin2021nerf} and distractors or by guiding active mapping~\cite{zhan2022activermap,lee2022uncertainty} and next‐view selection~\cite{fisherrf,ran2023neurar} for autonomous reconstruction, rather than addressing uncertainty for camera localization.
Moreover, since most formulations depend on the query-point representation inherent to NeRF~\cite{mildenhall2021nerf}, they are not directly compatible with recent explicit models such as 3D Gaussian Splatting~\cite{kerbl20233d}.
In contrast, our work investigates how geometric and pose uncertainties jointly affect localization reliability and introduces a probabilistic refinement framework that explicitly incorporates these uncertainties into the localization process.

%% file: sec/3_method.tex
\section{Method}
\subsection{Overview}
\label{sec:sec3.1}

\begin{figure*}[!ht]
  \centering
  \includegraphics[width=1\textwidth]{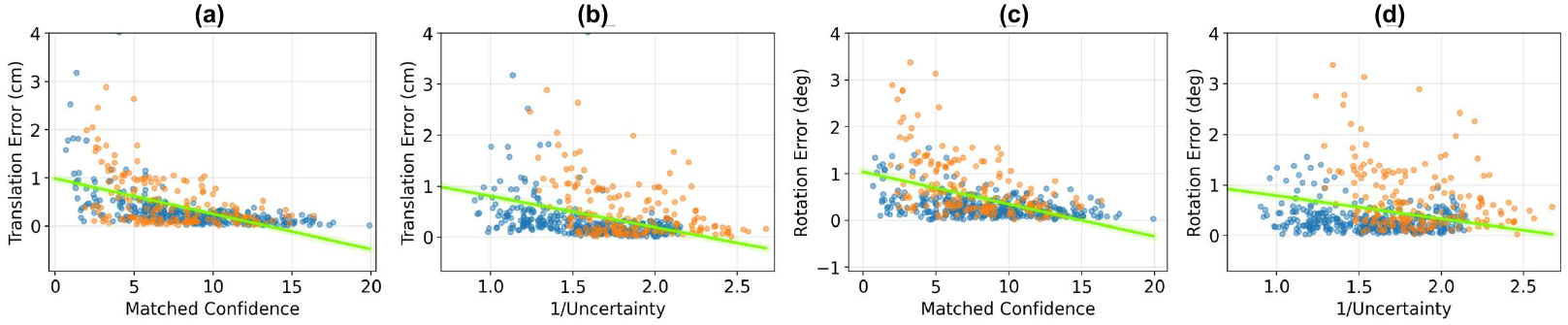}
\caption{\textbf{Distributions of Error–Confidence and Error–Uncertainty.} We visualize the relationship between translation error and the aggregated matcher confidence (a,c) or aggregated depth uncertainty (b,d) for frames from the Kings (blue) and Hospital (orange) scenes of the Cambridge dataset. For each frame, the x-axis value represents the sum of confidence or uncertainty across all its 2D–2D correspondences. The green trend line illustrates that higher confidence and lower uncertainty consistently correlate with smaller localization errors.}
\label{fig:distribution}
\vspace{-0.3cm}
\end{figure*}

\paragraph{Pose Refinement on 3DGS.}
In this work, we consider the uncertainty in camera pose refinement on 3DGS and propose a probabilistic pose refinement framework, as illustrated in \cref{fig:compare}. Pose refinement utilizing Gaussian Splatting~\cite{liu2024gs,huang2025sparse,li2024unleashing} typically rely on deterministic pipelines, where the pose estimation module predicts the pose prior $\mathbf{T}_\text{prior} = [\mathbf{R}_i, \mathbf{t}_i] \in SE(3)$, which is then refined through Gaussian Splatting $\mathcal{G}$. 
In these pipelines, a rendered image $I_r$ from estimated pose $\mathbf{T}_\text{prior}$ and query image $I_q$ are then used to extract 2D-2D correspondences $C_{q,r}$ by leveraging 2D matching modules $\mathcal{M}$ like MASt3r~\cite{leroy2024grounding} and LoFTR~\cite{sun2021loftr}. These 2D-2D pixel correspondences $C_{q,r} = \{ (\mathbf{q}_i, \mathbf{r}_i) \}$ are subsequently extended to 2D-3D correspondences $\mathcal{C} = \{ (\mathbf{q}_i, \mathbf{X}_i) \}$ using the rendered depth map $D_r$ and the camera intrinsics $K$:
\begin{equation}
    z_i = D_r(\mathbf{r}_i), \quad 
    \mathbf{X}_i = \mathbf{T}_\text{prior}
    \bigl( z_i K^{-1} \tilde{\mathbf{r}}_i \bigr),
\label{eq:lift}
\end{equation}
where $\tilde{\mathbf{r}}_i = [u_i, v_i, 1]^\top, \mathbf{r}_i = (u_i, v_i)$ denotes the pixel in the rendered image,
Finally, the pose prior $\mathbf{T}_\text{prior}$ is refined using a PnP-RANSAC solver~\cite{gao2003complete,lepetit2009ep,fischler1981random} to predict the camera pose $\mathbf{T}_\text{refined}$.

\paragraph{Pose Prior Uncertainty.} 
3DGS-based pose refinement pipeline does not require any additional training and is both simple and efficient, which is why it is adopted in many recent works. However, this pipeline makes two key assumptions: an accurate pose prior and the precise geometry of the scene. First, if the pose prior has a significant error or is located in an area with high occlusion within the scene, matching with the query image fails, or the matching quality may degrade. As shown in \cref{fig:compare}, the refined pose generated from such deterministic refinement is highly sensitive to the initial pose prior from the pose estimator~\cite{chen2022dfnet,chen2024map,brachmann2023accelerated}. This issue arises from the reliance on a single deterministic pose estimate. To address this, we utilize modified Monte Carlo Localization~\cite{kwon2007particle,thrun2005probabilistic}, which helps resolve such ambiguities and uncertainties by sampling multiple pose hypotheses and refining them probabilistically.

\paragraph{Geometric Uncertainty.} The 3DGS-based pose refinement pipeline is also dependent on the geometric quality of the 3DGS representation $\mathcal{G}$. In the existing pipeline, pose prior is refined via PnP-RANSAC with the correspondence between 3D coordinate map $X_r$ from \cref{eq:lift}. Even if the image synthesis is accurate, incorrect geometry can lead to erroneous 3D point information~\cite{goli2024bayes,fisherrf,yan2024cf}, which negatively impacts the final pose estimate. Building on these insights, we explore depth uncertainty in anchor-based Gaussian Splatting~\cite{lu2024scaffold}. Through the depth uncertainty estimated across the scene, we propose an uncertainty-guided point sampling strategy for pose refinement. This sampling scheme explicitly exploits geometric reliability to mitigate failure cases caused by erroneous 3D points during PnP optimization process, as detailed in \cref{sec:Monte Carlo Refinement}.

\subsection{Monte Carlo Refinement}
\label{sec:Monte Carlo Refinement}
To account for pose prior uncertainty, we adopt an MCL-inspired strategy that generates multiple pose hypotheses from the initial pose, refines each through our optimization procedure, and evaluates them via importance sampling.
MCL represents the camera pose as a set of weighted particles
$\mathcal{P} = \{ (\mathbf{T}^{(m)}, w^{(m)}) \}_{m=1}^{M}$,
where each $\mathbf{T}^{(m)} \in SE(3)$ denotes a sampled particle and $w^{(m)}$ its importance weight.
However, standard MCL frameworks are computationally inefficient for 6DoF pose in this high-dimensional state space. The traditional prediction phase, which typically involves simple random perturbation or sampling from a wide prior, is non-directed and requires a prohibitively large number of particles to adequately cover the $SE(3)$ manifold and converge. To address this, we replace prediction phase with local optimization as in \cref{sec:fisherrefinement}. This step guides each particle toward a nearby mode in the likelihood distribution. Crucially, this optimization acts as a directed pre-correction mechanism, significantly minimizing the uncertainty introduced by the wide prior.

After the pre-correction step, the importance weight $w^{(m)}$ of each particle $m$ needs to be identified to sample the most likely particle. Rather than applying photometric error~\cite{sun2023icomma,lin2023parallel,yen2021inerf}, which necessitates additional rendering and is highly susceptible to domain shift and scene appearance variations, we utilize two key metrics for efficient evaluation. Specifically, we leverage the matching confidence score $S_m$, which is a byproduct inherently generated by the 2D matching module $\mathcal{M}$, and the geometric uncertainty map $U_m$, described in \cref{sec:fisherrefinement}. As illustrated in \cref{fig:distribution}, a high matching confidence score $S_m$ and low geometric uncertainty $U_m$ strongly correlate with a reduced pose error, validating their use as reliable weighting factors. These metrics are then combined to quantify the fitness of the refined pose and determine whether the particle belongs to a high-likelihood region near the true camera pose.

Consequently, the importance weight $w^{(m)}$ for each refined particle $m$ is defined as the product of the two complementary factors, normalized over the particle set $\mathcal{P}$:
\begin{equation}
    w^{(m)} = \frac{\sum_{i=1}^{n_m} S_m(r_i) \cdot (1-U_m(r_i))}{\sum_{j=1}^{M} \sum_{i=1}^{n_j} S_j(r_i) \cdot (1-U_j(r_i))},
\end{equation}

where $r_i$ denotes the 2D-2D correspondence pixels on the rendered image. 
Following weight assignment, the final pose estimate is derived by resampling the particles based on $w^{(m)}$. The final estimated camera pose, $\mathbf{\hat{T}_\text{final}}$, is then typically extracted from the resulting particle set by either selecting the particle with the highest weight or by calculating the weighted mean of the particle poses in $SE(3)$ at the final iteration.
\begin{figure}[t]
  \centering
  \includegraphics[width=\columnwidth]{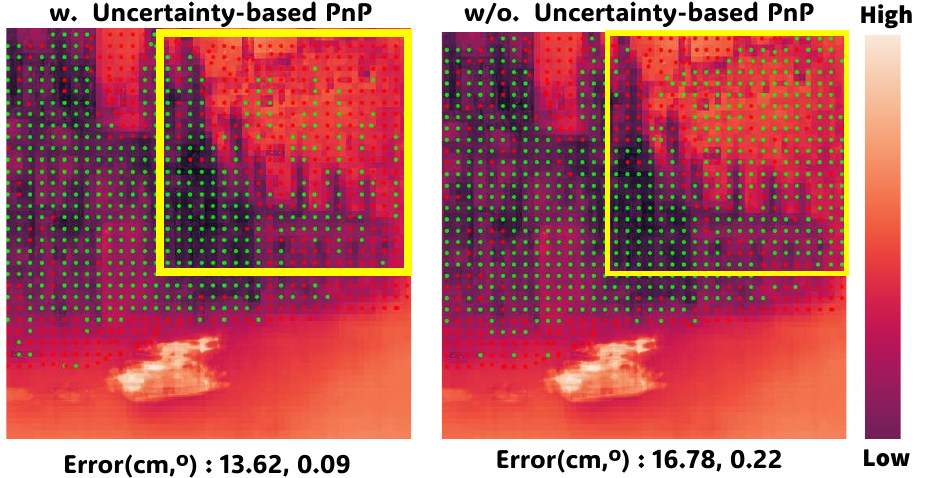}
\caption{\textbf{Effect of Uncertainty-based PnP.} We visualize the rendered uncertainty map together with 2D–2D correspondences projected onto the synthesized view. Green and red points denote inlier and outlier points, respectively. When applying our uncertainty-guided PnP, correspondences located in high-uncertainty regions are naturally suppressed, yielding more reliable inliers and improved pose estimate results.}
\label{fig:UPnP}
\vspace{-0.5cm}
\end{figure}

\subsection{Fisher Information Guided Pose Refinement}
\label{sec:fisherrefinement}
\paragraph{Fisher Information Approximation.}
To identify geometric uncertainty from 3DGS into pose refinement, we extend the Fisher Information–based depth uncertainty to an anchor-based Gaussian Splatting representation~\cite{lu2024scaffold} inspired by FisherRF~\cite{fisherrf}. 
Let the overall 3DGS model parameters be expressed as 
\begin{equation}
\theta = (\phi,\, g), \qquad g = D(\phi),
\end{equation}
where $\phi$ denotes the anchor parameters and $g$ the Gaussian attributes generated from decoder $D$. 
Following~\cite{fisherrf}, the Fisher Information 
can be approximated under standard regularity conditions~\cite{schervish2012theory} 
as the expected negative Hessian of the log-likelihood:
\begin{equation}
\mathcal{I}(x,\theta)
=-\,\mathbb{E}\!\left[\nabla^2_{\theta}
  \log p(\mathbf{y}\mid\mathbf{x},\theta)\right]
= \mathrm{H}''[y|x,\theta],
\end{equation}
where $x$ denotes the rendering pose. 
The Hessian matrix $\mathrm{H}''$ is computed from the Jacobian 
of the rendered outputs with respect to $\theta$, 
which can be efficiently obtained via a single backward pass 
for a given view. 
In practice, we apply a Laplace approximation~\cite{daxberger2021laplace,mackay1992bayesian} that replaces the full Hessian 
with its diagonal elements and a small regularization term:
\begin{equation}
    \mathrm{H}''[y|x,\theta] \simeq 
    \mathrm{diag}\big(
    (\nabla_{\theta} f)^{\!\top}
    (\nabla_{\theta} f)\big)
    + \lambda I.
\end{equation}

\paragraph{Aggregating Depth Uncertainty.} To quantify the geometric uncertainty, we aggregate Fisher Information $\mathcal{I}(x_i,\theta)$ from the given training views $x_i \in X$ in a global matrix $\mathrm{G}$,
defined as $\mathrm{G} = \sum_{i=1}^{N_\text{views}} \mathrm{H}''[y|x_i,\theta]$.
Then, the uncertainty $U$ of the estimated view $x_q$ is computed using the 3DGS rendering formulation, where each Gaussian’s geometric uncertainty is weighted by its depth contribution along the ray. By normalizing over the total number of Gaussians $N$, this operation effectively projects the accumulated 3D uncertainty into a per-pixel 2D uncertainty map:

\begin{equation}
\label{eq:uncertainty}
 U = \frac{1}{N}\sum_{i=1}^{N} \operatorname{tr}(\mathrm{G}_i)\, d_i\alpha_i \prod\nolimits_{j=1}^{i-1}(1-\alpha_j).
\end{equation}

Here, $d_i$ and $\mathrm{G}_n$ represent the depth of gaussian $i$ and the submatrix of $\mathrm{G}$ 
corresponding to the $i$th Gaussian point seen from the rendering view.

\paragraph{Pose Estimation via Geometric Uncertainty.} To incorporate the Fisher-based geometric uncertainty $G$ into the pose optimization procedure, the 2D-3D correspondences $\mathcal{C} = \{ (\mathbf{q}_i, \mathbf{X}_i) \}$ from \cref{eq:lift} are associated with a specific geometric uncertainty $U(r_i)$. This $U(r_i)$ quantifies the local geometric reliability of the reconstructed point $\mathbf{X}_i$. Unlike the simple RANSAC-Sampler, which relies on uniform sampling, our framework introduces a probabilistic bias from uncertainty fields. We construct sampling weights $s_i$ for each correspondence $\mathcal{C}$ based on this geometric reliability:
\begin{equation}
    s_i = e^{-\beta \, \bar{U}(r_i)} + \epsilon,
\end{equation}
where $\bar{U}\in[0,1]$ denotes the normalized uncertainty with all the uncertainty values of the matched pixels.
This formulation biases the sampling stage towards geometrically reliable correspondences by assigning higher weights $s_i$. At each refinement iteration, the minimal 2D-3D subset $\mathcal{S}_k \subset \mathcal{C}$ is sampled according to the probabilistic weights $s_i$. The estimated pose $\mathbf{T}_k$ is then calculated using the EPnP algorithm~\cite{lepetit2009ep} iteratively. This iterative process ensures that each newly generated hypothesis is inherently guided toward stable geometric anchors, as illustrated in \cref{fig:UPnP}. Furthermore, we extend this bias to the consensus evaluation. The final pose $\mathbf{T}_\text{refined}$ is selected by maximizing the weighted consensus $\sum s_i$.

\begin{figure*}[!ht]
  \centering
  \includegraphics[width=1\textwidth]{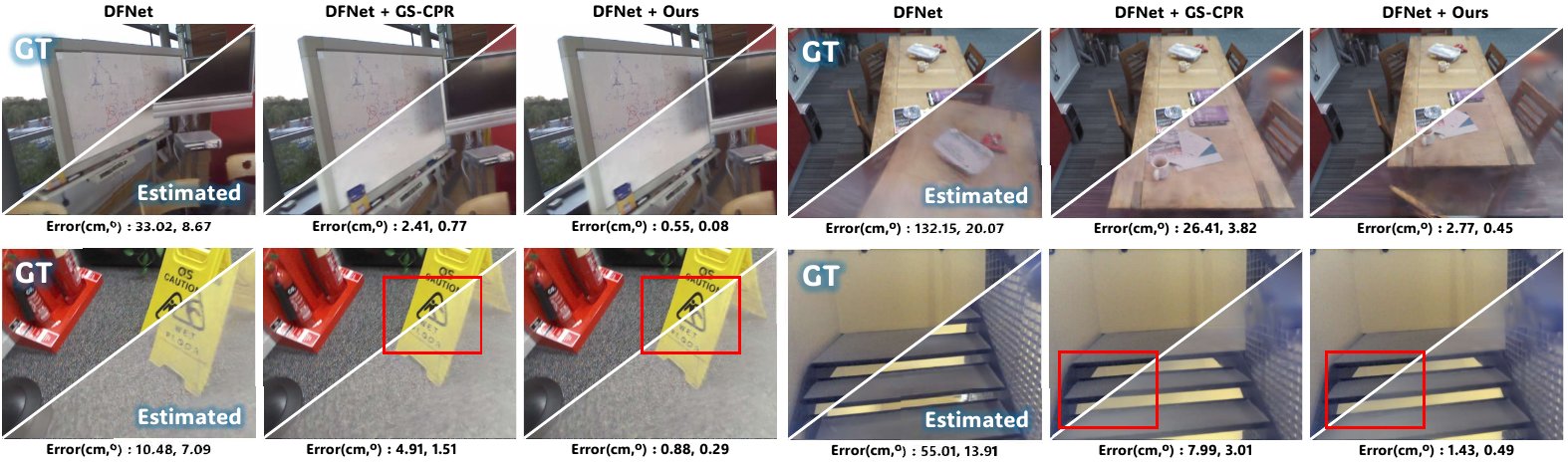}
\caption{\textbf{Visualization of Localization Quality on the 7-Scenes dataset.} Each triplet of images compares the ground-truth view (top-left) with the view rendered from (i) the input pose prior, (ii) the baseline GS-CPR refinement, and (iii) our UGS-Loc refinement (bottom-right).
A closer alignment along the diagonal boundary indicates better pose accuracy.}
\label{fig:result_figure}
\vspace{-0.2cm}
\end{figure*}

%% file: sec/4_exps.tex
\begin{table*}[!ht]
\caption{\textbf{Localization Results on 7Scenes dataset}. We report the median translation errors (cm) and rotation errors ($^\circ$). The best and second-best results are in \textbf{bold} and \underline{underline}.}
\centering
\setlength{\tabcolsep}{4pt}
\begin{threeparttable}
\resizebox{0.9\textwidth}{!}{
\begin{tabular}{l|c|ccccccc|c}
\toprule    & Methods & Chess  & Fire  & Heads & Office & Pumpkin  & Redkitchen & Stairs  & Avg. $\downarrow$ [$\text{cm}/^\circ$] \\
\midrule
 \multirow{5}{*}{\rotatebox{90}{APR}}& PoseNet~\cite{kendall2015posenet} & 10/4.02 & 27/10.0& 18/13.0 & 17/5.97& 19/4.67& 22/5.91 & 35/10.5 &  21/7.74\\
 & DFNet~\cite{chen2022dfnet}& 3/1.12 &6/2.30 & 4/2.29& 6/1.54
&7/1.92 & 7/1.74 & 12/2.63 & 6/1.93 \\
 & Marepo~\cite{chen2024map}& 1.9/0.83 &2.3/0.92 &2.1/1.24 & 2.9/0.93& 2.5/0.88
& 2.9/0.98 & 5.9/1.48 & 2.9/1.04 \\
& RAP~\cite{li2024unleashing} & 1/0.84 & 6/3.44 & 4/5.48 & 5/1.92 & 4/1.71 & 7/2.11 & 9/2.10 & 5/2.51 \\
 \midrule
\multirow{3}{*}{\rotatebox{90}{SCR}} & DSAC*~\cite{brachmann2017dsac}& 0.5/0.17 &0.8/0.28 & 0.5/0.34& 1.2/0.34&1.2/0.28 &  0.7/0.21& 2.7/0.78 & 1.1/0.34 \\
 & ACE~\cite{brachmann2023accelerated}& 0.5/0.18 &0.8/0.33 & 0.5/0.33&1.0/0.29&1.0/0.22 & 0.8/0.2 & 2.9/0.81 & 1.1/0.34 \\
 & GLACE~\cite{wang2024glace}& 0.6/0.18 & 0.9/0.34& 0.6/0.34& 1.1/0.29&0.9/0.23 &0.8/0.20  & 3.2/0.93 &  1.2/0.36\\
 \midrule
\multirow{11}{*}{\rotatebox{90}{NRP}}  &FQN-MN~\cite{germain2022feature} & 4.1/1.31 & 10.5/2.97 & 9.2/2.45 & 3.6/2.36 & 4.6/1.76& 16.1/4.42 & 139.5/34.67 & 28/7.3\\
&CrossFire~\cite{moreau2023crossfire} & 1/0.4 &5/1.9 &3/2.3 & 5/1.6 & 3/0.8& 2/0.8 &12/1.9  &4.4/1.38\\
& pNeRFLoc~\cite{zhao2024pnerfloc} & 2/0.8 & 2/0.88 & 1/0.83 & 3/1.05 & 6/1.51 & 5/1.54 & 32/5.73 & 7.3/1.76\\
&DFNet + NeFeS$_{50}$~\cite{chen2024neural} &2/0.57&2/0.74 &2/1.28 &2/0.56 &2/0.55 & 2/0.57 &5/1.28 & 2.4/0.79  \\
& NeRFMatch~\cite{zhou2024nerfect} & 0.9/0.3& 1.1/0.4& 1.5/1.0 &3.0/0.8 &2.2/0.6 &1.0/0.3 &10.1/1.7& 2.8/0.7 \\
&MCLoc~\cite{trivigno2024unreasonable}  &2/0.8 &3/1.4 &3/1.3& 4/1.3& 5/1.6& 6/1.6& 6/2.0 & 4.1/1.43\\
& DFNet + GS-CPR~\cite{liu2024gs}  & 0.7/0.20 & 0.9/0.32& 0.6/0.36 & 1.2/0.32 & 1.3/0.31  & 0.9/0.25 & 2.2/0.61 & 1.1/0.34\\
& Marepo + GS-CPR~\cite{liu2024gs} & 0.6/0.18 & 0.7/0.28 & 0.5/0.32 & 1.1/0.29& 1.0/0.26& 0.8/0.21 & 1.5/0.44 & 0.9/0.28\\
& ACE + GS-CPR~\cite{liu2024gs} & 0.5/\underline{0.15} & 0.6/0.25& \underline{0.4}/0.28& 0.9/0.26 & 1.0/ 0.23& 0.7/0.17 & 1.4/0.42 & 0.8/0.25 \\

& STDLoc~\cite{huang2025sparse} & 0.46/\underline{0.15} & 0.57/\underline{0.24} & 0.45/\underline{0.26} & 0.86/0.24 & 0.93/\underline{0.21} & 0.63/0.19 & 1.42/0.41 & 0.76/\underline{0.24} \\
& DFNet + \textbf{Ours} & \textbf{0.36}/0.16 & \underline{0.48}/\textbf{0.20} & \textbf{0.37}/\textbf{0.25} & \textbf{0.77}/\textbf{0.21} & \underline{0.81}/\textbf{0.18}   &  \underline{0.59}/\textbf{0.15} & 1.21/0.35 & 0.66/\textbf{0.21} \\
 & Marepo + \textbf{Ours} & \underline{0.37}/\textbf{0.12} & \underline{0.48}/\textbf{0.20} & \textbf{0.36}/\underline{0.26} & \underline{0.78}/\underline{0.22} & \underline{0.81}/\textbf{0.18}   &  \underline{0.59}/\textbf{0.15} & \underline{1.13}/\underline{0.34} & \underline{0.65}//\textbf{0.21}\\
& ACE + \textbf{Ours} & \underline{0.37}/\textbf{0.12} & \textbf{0.47}/\textbf{0.20}& \textbf{0.36}/\textbf{0.25}& \textbf{0.77}/\underline{0.22} & \textbf{0.79}/\textbf{0.18}& \textbf{0.58}/\textbf{0.15} & \textbf{1.11}/\textbf{0.33} & \textbf{0.64}/\textbf{0.21} \\

\bottomrule
\end{tabular}
}
\end{threeparttable}
\label{tab:acc_7s}
\end{table*}

\section{Experiments}
In this section, we present the localization performance and analysis of our proposed pose refinement pipeline.
\cref{subsec:loc_result} reports localization accuracy across both indoor and outdoor benchmarks.
\cref{subsec:Ablation} provides a detailed ablation study, examining the key components of our framework, including the comparison against simple iterative refinement, the influence of particle count, and the impact of different matching modules.

\vspace{-0.1cm}
\paragraph{Datasets.} We evaluate our method on three widely used visual localization benchmarks: 7Scenes~\cite{shotton2013scene}, 12Scenes~\cite{valentin2016learning}, and Cambridge Landmarks~\cite{kendall2015posenet}.
The 7Scenes dataset~\cite{shotton2013scene} consists of seven indoor environments, from offices to living rooms, with spatial volumes ranging from approximately 1 m³ to 18 m³.
The 12Scenes~\cite{valentin2016learning} dataset extends this setting to larger indoor environments (14 m³ – 79 m³), providing more complex scene layouts and viewpoint diversity.
The Cambridge Landmarks~\cite{kendall2015posenet} dataset contains four large-scale outdoor scenes with spatial extents from 875 m² to 5600 m², capturing various urban landmarks under significant illumination and viewpoint changes. 

\paragraph{Implementation Details.}
In Monte Carlo Refinement, although lightweight matching~\cite{lindenberger2023lightglue,sarlin2020superglue,detone2018superpoint} modules are employed, increasing the number of iterations $\mathcal{N}$ or particles $m$ leads to a substantial rise in inference time, due to the repeated matching and pose–optimization steps. Therefore, we maintain the entire pose refinement process to $\mathcal{N}=2$ with $m=8$ particles throughout all experiments. To further improve efficiency, the pose refinement and weighting of each particle are processed in parallel using multi-processing as \cite{trivigno2024unreasonable}. For the benchmark experiments, we use MASt3r~\cite{leroy2024grounding} as the matching module without finetuning, while additional studies on alternative matching modules are presented in \cref{subsec:Ablation} and supplementary material. For the perturbation noise of Monte Carlo Refinement, the first iteration samples translation and rotation perturbations from a uniform distribution with ranges of 10cm and 0.01$^\circ$, respectively.
while subsequent iterations use 1\,cm translation and 0.01$^\circ$ rotation noise. To derive the Fisher Information $\mathcal{I}(\theta)$ on Scaffold-GS parameters~\cite{lu2024scaffold}, only the anchor feature and the local Gaussian offset is explored as the parameters $\theta$. All experiments are conducted on an NVIDIA RTX A5000 GPU. Following GS-CPR~\cite{liu2024gs}, we train the Scaffold-GS~\cite{lu2024scaffold} model for 30k iterations and adopt the exposure adaptation strategy. With our standard localization configuration ($m=8$), rendering all particle views requires 190ms, and the batchified MASt3r inference takes 492ms. The uncertainty-guided PnP optimization with MASt3r matcher, executed in a multi-processing setting, adds 377ms. Overall, the end-to-end inference time is 1.1s per iteration. The total runtime can be adjusted by varying the number of particles. Notably, although MCLoc adopts a similar Monte Carlo–based refinement pipeline, its 80 iterations procedure requires around 2.4s per query, while our method remains within a practical runtime range under comparable settings.

\paragraph{Baselines.}
We compare its localization performance with state-of-the-art approaches across three benchmark datasets~\cite{shotton2013scene, lin2024learning, kendall2015posenet}. The comparison includes absolute pose regression (APR) method~\cite{shavit2021learning}, including PoseNet~\cite{kendall2015posenet}, DFNet~\cite{chen2022dfnet}, LENS~\cite{moreau2022lens}, and RAP~\cite{li2024unleashing};
scene coordinate regression (SCR) methods such as DSAC*~\cite{brachmann2017dsac}, ACE~\cite{brachmann2023accelerated}, and GLACE~\cite{wang2024glace};
and neural rendering-based post-refinement (NRP) methods~\cite{liu2024gs,moreau2023crossfire,trivigno2024unreasonable,huang2025sparse, germain2022feature, chen2024neural, zhou2024nerfect,zhao2024pnerfloc},
which leverage 2D-3D correspondence derived from neural scene representation or refine the initial pose by synthesising a novel view.

\subsection{Benchmark Results}
\label{subsec:loc_result}

\begin{table}[t]
\caption{\textbf{Average Recall on 7-Scenes.} We report the recall rate within the $[2\text{cm}, 2^\circ]$ and $[5\text{cm}, 5^\circ]$ error thresholds across the 7Scenes. } 
\centering
\setlength{\tabcolsep}{4pt}
\begin{threeparttable}
\resizebox{\columnwidth}{!}{
\begin{tabular}{c c c}

\toprule     
Datasets & \multicolumn{2}{c}{7scenes}\\
\cmidrule(r){1-1} \cmidrule(r){2-3} 
Methods & Avg. Acc $\uparrow$ [$5\text{cm},5^\circ$] &Avg. Acc $\uparrow$ [$2\text{cm},2^\circ$] 
\\\cmidrule(r){1-1} \cmidrule(r){2-2} \cmidrule(r){3-3}
 DFNet& 43.1 &  8.4  \\
 DFNet + GS-CPR & 94.2 & 76.5 \\
DFNet + GS-CPR$^2$ & 98.4 & 86.1 \\
 DFNet + \textbf{UGS-Loc} & \textbf{99.7} &  \textbf{95.1} 
 \\\cmidrule(r){1-1} \cmidrule(r){2-2} \cmidrule(r){3-3} 
 ACE& 97.1 &  83.3 \\
 ACE + GS-CPR & \textbf{100} & 93.1 \\
 ACE + GS-CPR$^2$ & \textbf{100} & 93.1 \\
  ACE + \textbf{UGS-Loc} & \textbf{100} &  \textbf{95.6} \\
\bottomrule
\end{tabular}}
\end{threeparttable}
\label{tab:7s_acc_percent}
\vspace{-0.2cm}
\end{table}

\noindent\textbf{7-scenes~\cite{glocker2013real,shotton2013scene}.}
As shown in \cref{tab:acc_7s}, we present the median translation and rotation errors for each scene. Our proposed method, UGS-Loc, achieves approximately 40\% improvement from the prior poses estimated from ACE~\cite{brachmann2023accelerated}, DFNet~\cite{chen2022dfnet}, and merapo~\cite{chen2024map} in terms of both median translation and rotation errors. Furthermore, when compared to GS-CPR~\cite{liu2024gs}, a purely deterministic radiance field–based pose refinement framework, our method shows an additional improvement about 20\% for three initial pose estimators~\cite{brachmann2023accelerated,chen2022dfnet,chen2024map}, despite employing a similar refinement procedure. Also, we demonstrate average accuracy of a $2\text{cm},2^\circ$ and $5\text{cm},5^\circ$ psoe error threshold in \cref{tab:7s_acc_percent}. In \cref{tab:7s_acc_percent}, we additionally include the results of GS-CPR$^2$, which applies the refinement structure of GS-CPR in an iterative manner for a fair comparison. GS-CPR$^2$ performs relocalization iteratively by using the refined pose as the new initial pose. Our proposed UGS-Loc demonstrates higher accuracy than the original non-iterative refinement structure and also outperforms the iterative GS-CPR$^2$ by 10\% under the $2\text{cm}, 2^\circ$ threshold. This improvement stems not from iterative refinement but from effectively addressing the limitations of the deterministic framework through our proposed probabilistic formulation.

\begin{table}[t]
\caption{\textbf{Localization Results on 12Scenes dataset.} We present the average percentage of frames whose estimated poses satisfy the $[2\text{cm}, 2^\circ]$ and $[5\text{cm}, 5^\circ]$ thresholds, along with the median translation and rotation errors (cm/$^\circ$) across the 12Scenes. }
\centering
\newcolumntype{d}{D{.}{.}{2}}
\setlength{\tabcolsep}{4pt}
\begin{threeparttable}
\resizebox{\columnwidth}{!}{
\begin{tabular}{c c c c}
\toprule
Methods& Avg. Err $\downarrow$ [$\text{cm}/^\circ$] & Avg. $\uparrow$ [$5\text{cm},5^\circ$] & Avg. $\uparrow$ [$2\text{cm},2^\circ$]\\
\midrule
Marepo &2.1/1.04 & 95 & 50.4\\
DSAC* & 0.5/0.25 & 99.8  & 96.7 \\
ACE & 0.7/0.26 & \textbf{100} & 97.2\\
GLACE & 0.7/0.25 & \textbf{100} & 97.5 \\
Marepo + GS-CPR & 0.7/0.28  & 98.9 & 90.9 \\
Marepo + \textbf{UGS-Loc} & 0.5/0.21  & 99.4 & 95.3 \\
ACE + GS-CPR & 0.5/0.21  &\textbf{100} & 98.7\\
ACE + \textbf{UGS-Loc} & \textbf{0.4/0.18}  &\textbf{100} & \textbf{99.1}\\
\midrule
\end{tabular}
}
\end{threeparttable}
\label{tab:12s_scenes}
\vspace{-0.2cm}
\end{table}

\begin{table}[t]
\centering
\caption{\textbf{Localization Results on Cambridge Landmark dataset.} We report the median translation errors (cm) and rotation errors ($^\circ$). The best results are in \textbf{bold} and second-best results are indicated with an \underline{underline}.}
\setlength{\tabcolsep}{4pt} %
\begin{threeparttable}
\resizebox{\columnwidth}{!}{
\begin{tabular}{l|c|cccc|c}
\toprule
 &Methods & Kings  & Hospital  & Shop & Church  &Avg. $\downarrow$ [$\text{cm}/^\circ$]   \\\midrule
 \multirow{6}{*}{APR}&PoseNet~\cite{kendall2015posenet} & 93/2.73 & 224/7.88 & 147/6.62 & 237/5.94 & 175/5.79 \\
&MS-Transformer~\cite{shavit2021learning}  & 85/1.45 &175/2.43& 88/3.20 & 166/4.12 & 129/2.80 \\
&LENS~\cite{moreau2022lens} & 33/0.5 &44/0.9& 27/1.6 & 53/1.6 & 39/1.15 \\
&DFNet~\cite{chen2022dfnet} & 73/2.37 & 200/2.98 & 67/2.21 & 137/4.02 & 119/2.90 \\
& RAP~\cite{li2024unleashing} & 58/0.87& 81/1.20& 33/1.64& 60/1.83& 58/1.39\\
\midrule
\multirow{2}{*}{SCR}& ACE~\cite{brachmann2023accelerated} & 29/0.38 & 31/0.61 & 5/0.3 & 19/0.6 & 21/0.47 \\
&GLACE~\cite{wang2024glace} & 20/0.32 & 20/0.41 & 5/0.22 & 9/0.3 & 14/0.32\\
\midrule
\multirow{11}{*}{NRP}  &FQN-MN~\cite{germain2022feature} & 28/0.4& 54/0.8 & 13/0.6 & 58/2 & 38/1 \\
&CrossFire~\cite{moreau2023crossfire}& 47/0.7& 43/0.7 & 20/1.2 & 39/1.4 & 37/1 \\
&DFNet + NeFeS$_{30}$~\cite{chen2024neural}& 37/0.64& 98/1.61 & 17/0.60 & 42/1.38 & 49/1.06 \\
&DFNet + NeFeS$_{50}$~\cite{chen2024neural}& 37/0.54&52/0.88 &15/0.53& 37/1.14& 35/0.77 \\
&HR-APR~\cite{liu2024hr} & 36/0.58&53/0.89 & 13/0.51 & 38/1.16 & 35/0.78\\
&MCLoc~\cite{trivigno2024unreasonable} &31/0.42& 39/0.73& 12/0.45& 26/0.8& 27/0.6\\
&RAP$_\text{ref}$~\cite{li2024unleashing} & \underline{18}/0.38 &22/0.42 &8/0.39 &16/0.47 &16/0.42\\
&DFNet + GS-CPR~\cite{liu2024gs} & 23/0.32 &42/0.74&10/0.36&27/0.62& 
26/0.51\\
&ACE + GS-CPR~\cite{liu2024gs} & 20/0.29&21/0.40&5/0.24&13/0.40& 15/0.33\\
&DFNet + \textbf{Ours} & 18.7/\underline{0.19} & \underline{14.5}/\textbf{0.29} & \textbf{3.9}/\underline{0.15}& \textbf{5.5}/\textbf{0.17}& \underline{10.7}/\textbf{0.20}\\
&ACE + \textbf{Ours} & \textbf{17.8}/\textbf{0.18} & \textbf{13.8}/\underline{0.30} & \underline{4.2}/\textbf{0.16}& \underline{6.3}/\underline{0.20}& \textbf{10.5}/\underline{0.21}\\
\bottomrule 
\end{tabular}
}
\end{threeparttable}
\vspace{-0.3cm}
\label{tab:acc_cam}
\end{table}

\noindent\textbf{12-scenes~\cite{valentin2016learning}.} As shown in \cref{tab:12s_scenes}, we report the quantitative localization performance using Merapo~\cite{chen2024map} and ACE~\cite{brachmann2023accelerated} as pose priors.
Given the strong baseline accuracy of SCR-based methods~\cite{brachmann2021visual,wang2024glace} on the $[5\text{cm}, 5^\circ]$ threshold, we focus on the strict threshold $[2\text{cm}, 2^\circ]$, where our approach achieves the highest accuracy of 99.1\%.
Moreover, our method attains the lowest median translation and rotation errors across all scenes, outperforming the conventional refinement pipeline GS-CPR by delivering more precise and stable localization results.

\noindent\textbf{Cambridge Landmark~\cite{kendall2015posenet}.}
As shown in \cref{tab:acc_cam}, we evaluate localization performance using DFNet~\cite{chen2022dfnet} and ACE~\cite{brachmann2023accelerated} as pose priors.
UGS-Loc reduces the median translation error of the baseline GS-CPR~\cite{liu2024gs} by approximately 30\%, demonstrating strong refinement capability.
Notably, our framework converges reliably within only two iterations, consistently achieving high accuracy regardless of the quality of the initial pose prior. Although DFNet’s localization performance is significantly weaker than ACE, often differing by five times, our refinement module delivers similarly accurate results for both priors. This illustrates the ability of UGS-Loc to mitigate the uncertainty introduced by suboptimal pose priors.
Remarkably, even in challenging cases such as the Shop scene, UGS-Loc with DFNet prior surpasses the refinement results obtained using ACE, further validating the robustness of our approach to prior noise.

\subsection{Ablation Study}
\label{subsec:Ablation}

\begin{figure}[t!]
  \centering
  \includegraphics[width=\columnwidth]{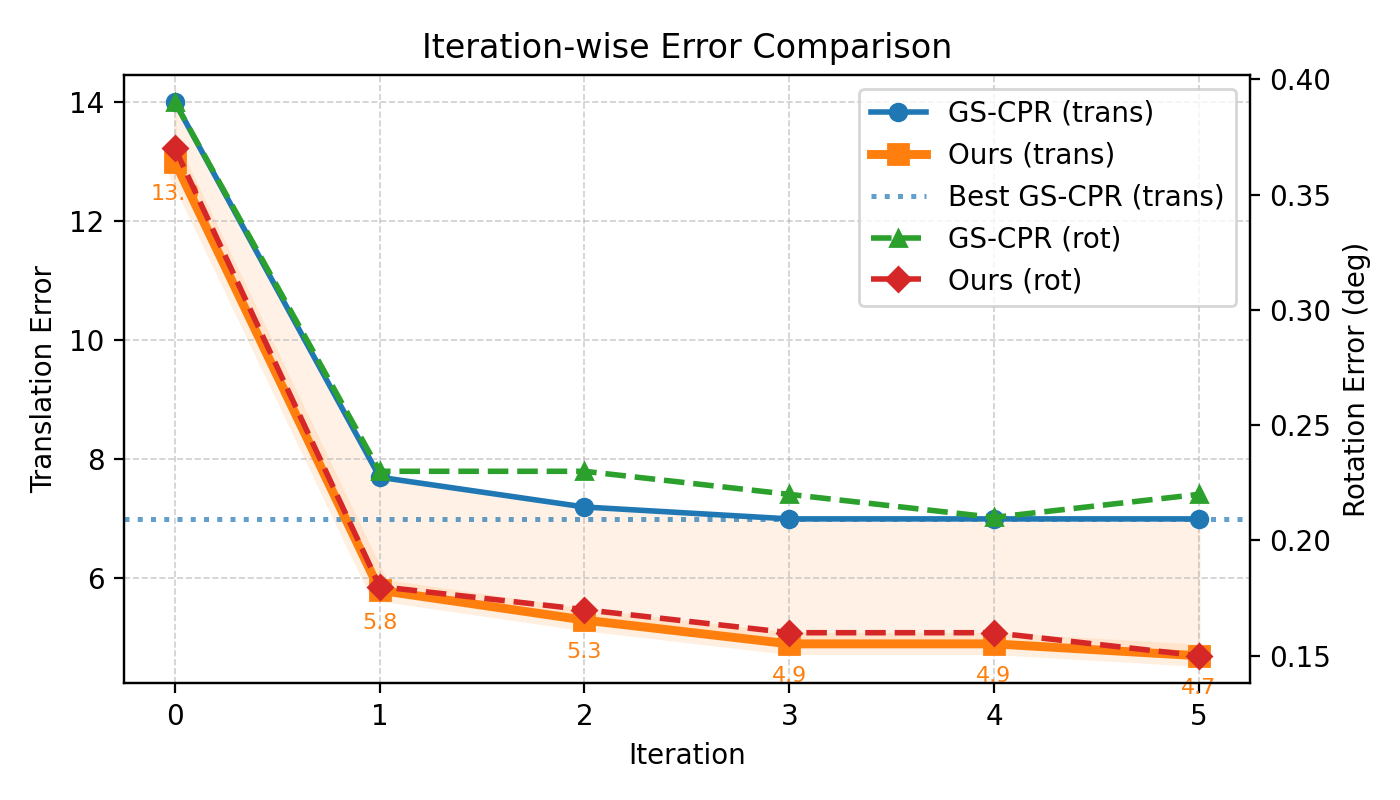}
\caption{\textbf{Comparison with Iteration-wise Pose Refinement.} We visualize median translation and rotation errors over iterations on Church scene of the Cambridge Landmarks, comparing our UGS-Loc with the iterative refinement strategy adopted in GS-CPR~\cite{liu2024gs}.}
\label{fig:fig_abl1}
\vspace{-0.5cm}
\end{figure}

\paragraph{Comparison with Simple Iterative Refinement.} We investigate whether the performance gain of our UGS-Loc simply stems from multiple refinement iterations.
To this end, we conduct an experiment on the Cambridge Landmarks dataset by iteratively applying the deterministic baseline pipeline~\cite{liu2024gs} several times and comparing it with our proposed probabilistic refinement.
As shown in \cref{fig:fig_abl1}, even though iterative refinement with GS-CPR gradually reduces translation and rotation errors, it quickly saturates after the first iteration. In contrast, our method achieves both consistent convergence and lower final errors, demonstrating that the improvement originates from our framework rather than from repeated deterministic optimization.

\begin{table}[t]
\centering
\caption{\textbf{Evaluating the Number of Particles for Refinement.} We report the median translation and rotation errors (cm/$^\circ$) on the Cambridge Landmarks dataset across different particle counts used in the Monte Carlo Refinement. PP and NP denote pose prior and the number of particles}
\setlength{\tabcolsep}{4pt} %
\begin{threeparttable}
\resizebox{\columnwidth}{!}{
\begin{tabular}{l|c|cccc|c}
\toprule
PP & NP & Kings  & Hospital  & Shop & Church  &Avg. $\downarrow$ [$\text{cm}/^\circ$]   \\\midrule

\multirow{4}{*}{DFNet} & 2 & 18.7/0.20 & 16.24/0.36& 5.5/0.20 & 6.6/0.20& 11.8/0.24\\
& 4 & 19.1/0.19 & 15.1/0.29 & 4.6/0.19 & 6.0/0.18& 
11.2/0.21\\
& 8 & 18.7/\textbf{0.19} & 14.50.29 & \textbf{3.9}/\textbf{0.15}& 5.5/0.17& 10.7/\textbf{0.20}\\
& 16 & \textbf{18.2}/0.20 & \textbf{13.8}/\textbf{0.27} & 4.2/0.16& \textbf{5.1}/\textbf{0.16}& \textbf{10.3}/\textbf{0.20}\\
\midrule

\multirow{4}{*}{ACE} & 2 & 17.8/0.21 & 14.8/0.30& 4.0/0.18 & 6.9/0.23& 
10.9/0.51\\
& 4 & 18.8/0.19 & 14.3/0.28 & 3.7/0.16 & 6.8/0.21& 
10.7/0.21\\
& 8 & 17.8/\textbf{0.18} & 13.8/0.30 & 4.2/0.16 & 6.3/0.20& 10.5/0.21\\
& 16 & \textbf{17.0}/\textbf{0.18} & \textbf{11.9}/\textbf{0.28} & \textbf{3.4}/\textbf{0.15}& \textbf{6.1}/\textbf{0.19}& \textbf{9.6}/\textbf{0.20}\\

\bottomrule 
\end{tabular}
}
\end{threeparttable}

\label{tab:num_abl}
\end{table}
\vspace{-0.3cm}
\paragraph{Effect of Particle Count on Pose Refinement.}
We analyze the effect of the number of particles used in the Monte Carlo Refinement. As shown in \cref{tab:num_abl}, we vary the number of particles from 2 to 16 while keeping the number of refinement iterations as 2. We observe that increasing the number of particles consistently improves both translation and rotation accuracy across all scenes. In particular, the performance gain is most pronounced when increasing the particle count. This suggests that using multiple pose hypotheses helps better capture the uncertainty of the pose prior and avoids convergence to suboptimal local minima.

\begin{table}[t]
\footnotesize
\centering
\caption{\textbf{Evaluation with Different 2D Matching Module.}
We compare our probabilistic refinement framework using sparse feature matchers (SuperPoint (SP)~\cite{detone2018superpoint} + LightGlue (LG)~\cite{lindenberger2023lightglue}) and dense correspondence model (MASt3r~\cite{leroy2024grounding}).}
\setlength{\tabcolsep}{4pt} %
\begin{threeparttable}
\resizebox{\columnwidth}{!}{
\begin{tabular}{c|c|cccc|c}
\toprule
Prior & Matchers & Kings  & Hospital  & Shop & Church  &Avg. $\downarrow$ [$\text{cm}/^\circ$]   \\\midrule
 \multicolumn{2}{c|}{Baseline} & 23/0.32 &42/0.74&10/0.36&27/0.62& 
26/0.51\\\midrule
 DFNet & SP + LG & 20/0.22 & 17/0.31 & 5.5/0.26& 7.7/0.25& 13/0.26 \\
 DFNet & MASt3r & 19/\textbf{0.19} & 15/0.32 & 4.3/0.17& \textbf{5.8}/\textbf{0.18}& 11/0.22 \\
 ACE & SP + LG & 19/0.23 & 15/\textbf{0.27} & \textbf{4.0}/0.19& 7.8/0.26& 11/0.26 \\
 ACE & MASt3r & \textbf{18}/0.20 & \textbf{14} /0.30 & 4.5/\textbf{0.16}& 6.5/0.21& 11/0.22 \\

\bottomrule 
\end{tabular}
}
\end{threeparttable}

\label{tab:abl_matching}
\end{table}

\paragraph{Various Matching Modules.} 
As shown in \cref{tab:abl_matching}, our method demonstrates localization results compared to a lightweight sparse feature matcher (SuperPoint~\cite{detone2018superpoint} + LightGlue~\cite{lindenberger2023lightglue}).
While MASt3r tends to yield slightly higher accuracy due to its strong context reasoning and dense feature aggregation, our uncertainty-aware refinement achieves comparable localization accuracy even with lightweight sparse matchers. We report GS-CPR with MASt3r as baseline.

%% file: sec/5_conclusion.tex
\section{Conclusion}
In this paper, we investigated the overlooked impact of prior pose and geometry uncertainty in 3D Gaussian Splatting-based pose refinement. To alleviate these issues, we proposed a probabilistic relocalization framework that combines Monte Carlo sampling with Fisher Information–guided PnP optimization. This design handles uncertainty without requiring any retraining or scene-specific adjustments. Extensive experiments across indoor and outdoor benchmarks demonstrate that our method consistently improves localization accuracy and remains robust regardless of the choice of pose prior or matching module. Overall, our findings highlight that incorporating uncertainty into the 3DGS pipeline yields a more stable and reliable pose refinement process.

%% file: sec/6_acknowledgement.tex
\section*{Acknowledgements}
This work was supported by Korea Evaluation Institute Of Industrial Technology (KEIT) grant funded by the Korea government(MOTIE) (No.20023455, Development of Cooperate Mapping, Environment Recognition and Autonomous Driving Technology for Multi Mobile Robots Operating in Large-scale Indoor Workspace), the KIST Institutional Program (No.2E33801-25-015) and the Institute of Information \& Communications Technology Planning \& Evaluation(IITP) grant funded by the Korea government(MSIT) (No.RS-2025-02219317, AI Star Fellowship, Kookmin University).

%% file: sec/X_suppl.tex
\maketitlesupplementary

\renewcommand\thesection{\Alph{section}} 
\renewcommand\thesubsection{\thesection.\arabic{subsection}} 
\renewcommand\thefigure{\Alph{figure}} 
\renewcommand\thetable{\Alph{table}} 

\crefname{section}{Sec.}{Secs.}
\Crefname{section}{Section}{Sections}
\Crefname{table}{Table}{Tables}
\crefname{table}{Tab.}{Tabs.}

\newcommand{\tabnohref}[1]{Tab.~{\color{red}#1}} 
\newcommand{\fignohref}[1]{Fig.~{\color{red}#1}} 
\newcommand{\secnohref}[1]{Sec.~{\color{red}#1}} 
\newcommand{\cnohref}[1]{[{\color{green}#1}]} 
\newcommand{\linenohref}[1]{Line~{\color{red}#1}}

\section*{Supplementary}

In the supplementary material, we show
\begin{itemize}
    \item integrating retrieval-based pose initialization into our method;
    \item a comparison using various matching methods;
    \item an ablation study analyzing each module in our pipeline;
    \item a runtime discussion;
    \item an additional implementation detail.

\end{itemize}

\section{Leveraging Image Retrieval Prior}

Pose refinement in 3D Gaussian Splatting relies heavily on the quality of the initial pose prior, since the refinement module only adjusts the pose locally around this estimate. Consequently, the choice of pose estimator providing the prior, typically an image retrieval system, APR, or SCR, plays a crucial role in determining the overall localization performance.
Among these options, image retrieval offers a unique advantage: unlike APR or SCR, it does not require any per-scene training. When combined with a 3DGS scene representation, this enables a purely geometry-driven localization pipeline that performs relocalization without additional learning.

\begin{figure}[!t]
  \centering
  \includegraphics[width=\columnwidth]{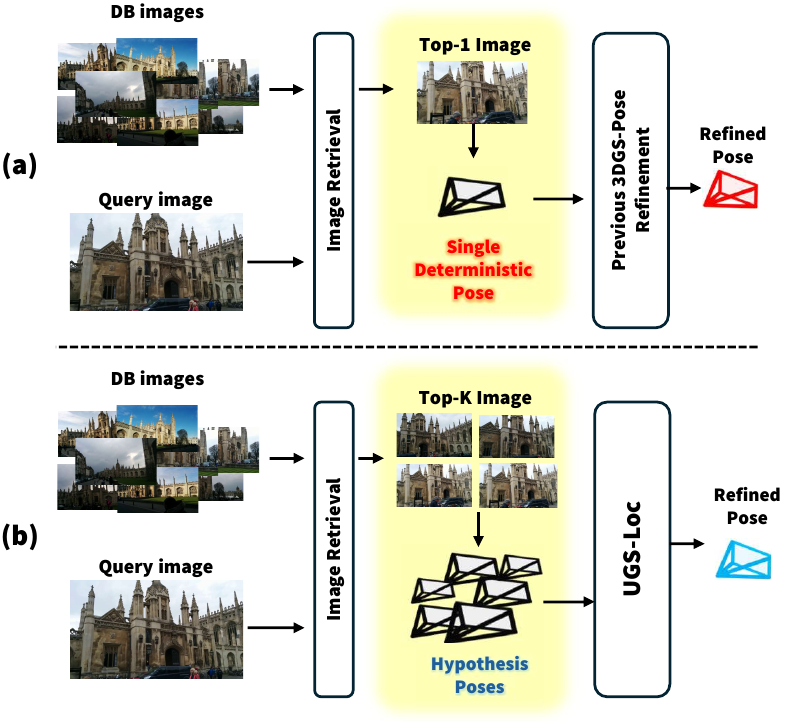}
\caption{\textbf{Leveraging Image Retrieval into UGS-Loc.}
Illustration comparing deterministic 3DGS-based pose refinement and our UGS-Loc when using image retrieval as a pose prior.
(a) Prior methods rely solely on the Top-1 retrieved image to obtain a single deterministic pose hypothesis, which is then refined through a 3DGS-based pipeline, making them vulnerable to incorrect or ambiguous retrieval results.
(b) Our UGS-Loc instead leverages Top-K retrieved images to generate multiple pose hypotheses, followed by importance weighting and refinement. This enables the localization to suppress misleading hypotheses and converge reliably toward the true camera pose.}
\label{fig:supp_IR_compare}
\end{figure}

\begin{figure}[!t]
  \centering
  \includegraphics[width=\columnwidth]{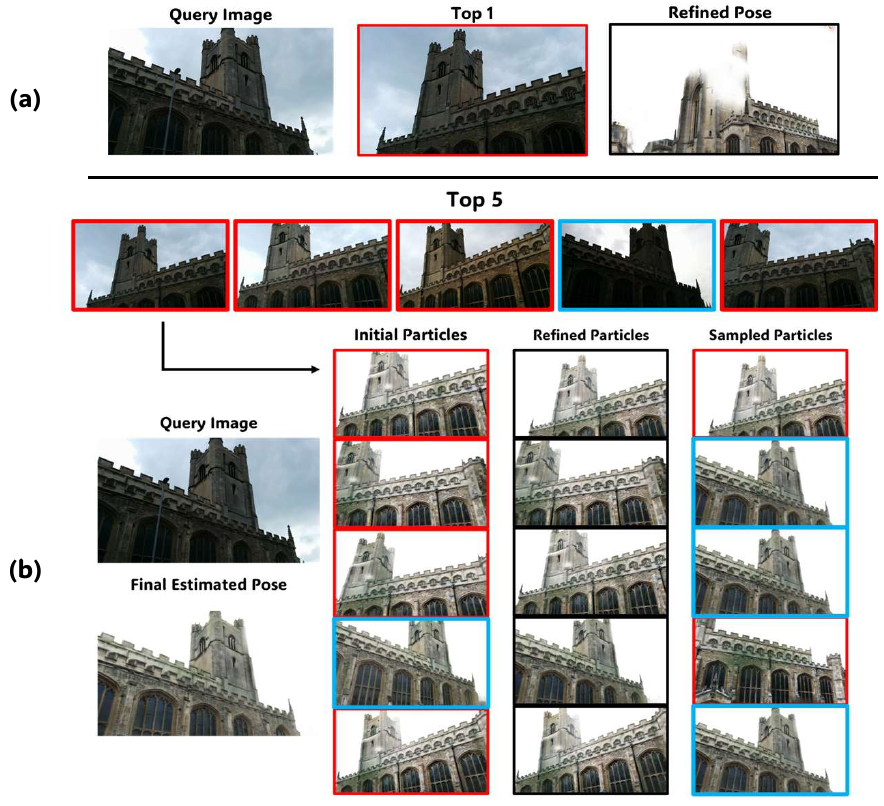}
\caption{\textbf{Visualization of UGS-Loc with Image Retrieval.}
We illustrate how integrating image retrieval with our pipeline enables robust pose refinement. In the Cambridge \textit{Church} scene, the deterministic Top-1 baseline (a) suffers from a failure case where the retrieved image (red box) views the church from the opposite side of the scene, yielding an incorrect pose prior. In contrast, using Top-K retrieved poses (b), UGS-Loc effectively suppresses such erroneous hypotheses and converges to the correct pose, highlighted by blue boxes.}
\label{fig:supp_IR_2}
\end{figure}

As illustrated in \cref{fig:supp_IR_compare}, conventional 3DGS-based refinement pipelines use the pose of the Top-1 retrieved database (DB) image as the pose prior. While this approach is simple and efficient, it inherently assumes that the retrieved Top-1 image is spatially close to the query. In practice, this assumption frequently breaks down.
Even if the retrieved DB image is visually similar to the query, it may still be captured from a different viewpoint, for example, the opposite side of the same building. ~\cref{fig:supp_IR_2} demonstrates such a failure case in the Cambridge Landmarks \textit{Church} scene. The Top-1 retrieved image exhibits high visual similarity, yet its pose lies on the opposite side of the church, leading to a large spatial discrepancy. Refining such an incorrect prior remains challenging, regardless of the rendering quality or scene detail available in the 3DGS model.

Our proposed UGS-Loc addresses this limitation by integrating retrieval results into the Monte Carlo refinement framework. Instead of relying on a single deterministic prior, we use the poses of the Top-$K$ retrieved DB images as initial hypotheses (particles). Each hypothesis is assigned an importance weight derived from matching confidence and geometric uncertainty, allowing the system to down-weight misleading candidates and preserve only the reliable ones through resampling.
As shown in ~\cref{fig:supp_IR_2}, although most initial particles lie on the incorrect side of the building, UGS-Loc samples the poses on the correct region by leveraging importance weighting, which leverage the confidence from the matching module and uncertainty of matching points. This demonstrates the robustness of our retrieval-integrated sampling scheme and highlights its advantage over traditional deterministic pose refinement.

\begin{table*}[ht!]
\scriptsize
\centering
\caption{\textbf{Localization Result on 7scenes with Image Retrieval.} We report the median translation error (cm) and rotation errors ($^\circ$). Pose prior is initialized with the poses of the database images retrieved by NetVlad~\cite{arandjelovic2016netvlad}. 
}
\setlength{\tabcolsep}{4pt} %
\begin{threeparttable}
\resizebox{\textwidth}{!}
{
\begin{tabular}{ccccccccccc}
\toprule
 Iteration & Method &Top-K & Chess  & Fire  & Heads & Office & Pumpkin  & Redkitchen & Stairs  & Avg. $\downarrow$ [$\text{cm}/^\circ$] \\\midrule
 1 & \multirow{2}{*}{GS-CPR} & 1 & 1.66/0.43 & 2.38/0.61 & 1.34/0.67 & 3.33/0.65 & 2.16/0.45 & 2.56/0.54& 4.51/1.10 & 2.56/0.64 \\
 2 &  & 1 & 0.56/0.16& 0.80/0.28 & 0.48/0.32 & 1.18/0.31 & 1.13/0.25& 0.77/0.19 & 2.24/0.66 & 1.03/0.31 \\
  \midrule
 2 & \textbf{UGS-Loc (Ours)} & 5 & \textbf{0.40}/\textbf{0.13}& \textbf{0.54}/\textbf{0.21} & \textbf{0.40}/\textbf{0.27} & \textbf{0.88}/\textbf{0.23} & \textbf{0.85}/\textbf{0.19}& \textbf{0.64}/\textbf{0.16} & \textbf{1.80}/\textbf{0.50}& \textbf{0.79}/\textbf{0.24} \\

\bottomrule 
\end{tabular}
}
\end{threeparttable}

\label{tab:supp_IR_7scenes}
\end{table*}

\begin{table}[ht!]
\scriptsize
\centering
\caption{\textbf{Average Recall on 7-Scenes with Image Retrieval.} We report the recall rate within the [$2\text{cm},2^\circ$] and [$5\text{cm},5^\circ$] error thresholds across the
7Scenes.}

\begin{threeparttable}
\resizebox{\columnwidth}{!}
{
\begin{tabular}{ccccc}
\toprule
 Iteration & Method &Top-K & Acc $\uparrow$ [$2\text{cm},2^\circ$] & Acc $\uparrow$ [$5\text{cm},5^\circ$] \\\midrule
 1 & \multirow{2}{*}{GS-CPR} & 1 & 43.0 & 76.6\\
 2 &  & 1 & 79.9 & 94.4 \\
  \midrule
 2 & \textbf{UGS-Loc} & 5 & \textbf{88.9} & \textbf{97.2} \\
  \midrule
\multicolumn{3}{c}{ACE} & 83.3 & 97.1 \\
\bottomrule 
\end{tabular}
}
\end{threeparttable}

\label{tab:supp_IR_7scenes_acc}
\end{table}

We evaluate the effectiveness of integrating image retrieval priors into our UGS-Loc framework, as shown in \cref{tab:supp_IR_7scenes,tab:supp_IR_camb}. We utilize NetVlad~\cite{arandjelovic2016netvlad} as a 
image retrieval model. For a fair comparison, we also include an iterative extension of the standard 3DGS-based pose refinement pipeline, in which the refined pose is repeatedly fed back as the new initialization. This allows us to directly compare UGS-Loc against both the conventional single-step refinement and its iterative variant.

\cref{tab:supp_IR_camb} reports the localization results on the large-scale Cambridge Landmarks dataset. Remarkably, when combined with image retrieval, our UGS-Loc achieves performance comparable to the performance of ACE~\cite{brachmann2023accelerated}, a Scene Coordinate Regression method that relies on per-scene training. Furthermore, compared to using only the Top-1 retrieved image as a deterministic pose prior, UGS-Loc reduces the median translation error by 17\%, despite operating under the same retrieval assumptions and without any additional learning.

Although the retrieval-based experiment involves one additional refinement iteration compared to the standard setting, the improvement remains noteworthy. This demonstrates that UGS-Loc effectively mitigates the uncertainty introduced by imperfect retrieval priors and consistently converges toward accurate poses. The evaluation on the 7Scenes dataset, presented in \cref{tab:supp_IR_7scenes}, further confirms this trend. UGS-Loc robustly handles pose priors obtained from retrieval, outperforming both the original 3DGS-based refinement and its iterative variant across all scenes.

\begin{table*}[ht!]
\scriptsize
\centering
\caption{\textbf{Localization Result on 7scenes with Image Retrieval.} We report the median translation error (cm) and rotation errors ($^\circ$). Pose prior is initialized with the poses of the database images retrieved by NetVlad~\cite{arandjelovic2016netvlad}. 
}
\setlength{\tabcolsep}{4pt} %
\begin{threeparttable}
\resizebox{0.8\textwidth}{!}
{
\begin{tabular}{cccccccc}
\toprule
 Iteration & Method &Top-K & Kings  & Hospital  & Shop & Church  &Avg. $\downarrow$ [$\text{cm}/^\circ$]   \\\midrule
 1 & \multirow{3}{*}{GS-CPR} & 1 & 26/0.33 & 35/0.57 & 13/0.43& 23/0.61& 24/0.49 \\
 2 &  & 1 & 24/0.26 & 27/0.46 & 6.2/0.25& 8.8/0.27& 17/0.31 \\
 3 &  & 1 & 19/0.23 & 20/0.41 & 5.9/0.21& 6.6/0.22 & 13/0.27 \\
 \midrule
 3 & \multirow{2}{*}{\textbf{UGS-Loc (Ours)}} & 5 & \textbf{18}/0.20 & \textbf{15}/0.32 &\textbf{ 4.3}/\textbf{0.17} &  5.4/0.18& \textbf{11}/0.22 \\
 3 &  & 10 & \textbf{18}/\textbf{0.19} & \textbf{15}/\textbf{0.31} & 4.5/\textbf{0.17} &\textbf{5.1}/\textbf{0.17}& \textbf{11}/\textbf{0.21} \\
 \midrule
  \multicolumn{3}{c}{ACE + UGS-Loc } & 18/0.20 & 14/0.30 & 4.5/0.16& 6.5/0.21& 11/0.22 \\

\bottomrule 
\end{tabular}
}
\end{threeparttable}

\label{tab:supp_IR_camb}
\end{table*}

\begin{figure*}[!t]
  \centering
  \includegraphics[width=\textwidth]{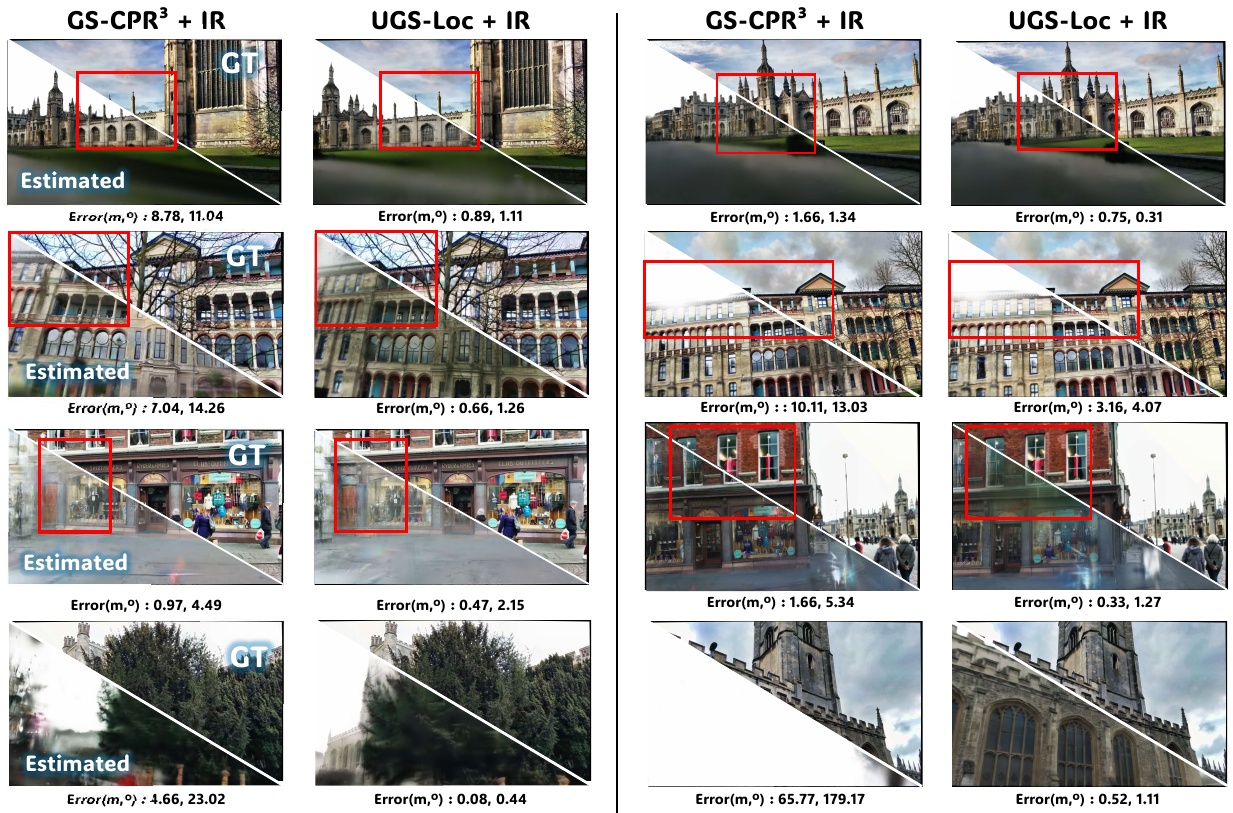}
\caption{\textbf{Visualization of Localization Quality on the Cambridge Landmark Dataset.}
Each pair of columns compares the ground-truth view (top-right) against the view rendered from (i) the iteratively modified GS-CPR refinement with 3 iterations (denoted as GS-CPR$^3$)~\cite{liu2024gs} and (ii) our UGS-Loc refinement (bottom-left). IR denotes that the pose prior is initialized with Image Retrieval (IR)~\cite{arandjelovic2016netvlad}. Tighter visual alignment along the diagonal boundary indicates a more accurate pose estimate. Red bounding boxes highlight regions where misalignment is most apparent, emphasizing how UGS-Loc corrects errors that remain unresolved by deterministic 3DGS-based refinement.}
\label{fig:supp_IR_camb}
\vspace{-0.2cm}
\end{figure*}

\section{Various Matching Modules}
\label{sec:supp_match}

Our UGS-Loc framework refines poses using 2D–3D correspondences obtained by lifting 2D–2D matches through rendered depth. Since correspondence quality directly depends on the underlying 2D matcher, we additionally evaluate UGS-Loc with a broader set of matching modules beyond the main paper, ELoFTR~\cite{wang2024efficient}, XFeat~\cite{potje2024xfeat}, and Match-Anything~\cite{li2024matching} , and report the median translation and rotation errors on the Cambridge Landmarks dataset in \cref{tab:supp_matcher}.

Across all matchers, UGS-Loc consistently achieves strong pose refinement performance, demonstrating that our method does not rely on any specific matching backbone. While transformer-based dense matchers generally provide higher correspondence coverage, even lightweight or classical matchers yield comparable localization accuracy when combined with UGS-Loc. This robustness highlights that the gains from our framework primarily arise from uncertainty-guided sampling and multi-hypothesis pose refinement, rather than from matcher capacity itself.

\begin{table*}[ht!]
\scriptsize
\centering
\caption{\textbf{Evaluation with Different 2D Matching Module.} We report the median translation error (cm) and rotation error ($^\circ$) on the Cambridge Landmark dataset~\cite{kendall2015posenet}. We utilize various matching modules to establish 2D-3D correspondences.
}
\setlength{\tabcolsep}{4pt} %
\begin{threeparttable}
\resizebox{0.7\textwidth}{!}
{
\begin{tabular}{ccccccc}
\toprule
Prior & Matchers & Kings  & Hospital  & Shop & Church  &Avg. $\downarrow$ [$\text{cm}/^\circ$]   \\\midrule
 \multirow{6}{*}{DFNet} & SP~\cite{sarlin2020superglue} + LG~\cite{lindenberger2023lightglue} & 20/0.22 & 17/0.31 & 5.5/0.26& 7.7/0.25& 13/0.26 \\
  & Xfeat~\cite{potje2024xfeat} & \textbf{19}/\textbf{0.18} & 15/0.28 & 5.5/0.26& 6.6/0.20& 12/0.23 \\
 & ELoFTR~\cite{wang2024efficient} & 20/0.21 & \textbf{13}/\textbf{0.25} & 4.1/0.19& 6.5/0.20 & \textbf{11}/\textbf{0.20} \\
 & MatchAny~\cite{li2024matching} & 20/0.20 & 14/\textbf{0.25} & \textbf{4.0}/\textbf{0.16}& 6.3/0.20 & \textbf{11}/\textbf{0.20} \\
 & MASt3r~\cite{leroy2024grounding} & \textbf{19}/0.19 & 15/0.30 & 4.3/0.17& \textbf{5.8}/\textbf{0.18}& \textbf{11}/0.22 \\
 \midrule
\multirow{6}{*}{ACE} & SP~\cite{detone2018superpoint} + LG~\cite{lindenberger2023lightglue} & 19/0.23 & 15/0.27 & 4.0/0.19& 7.8/0.26& \textbf{11}/0.26 \\
  & Xfeat~\cite{potje2024xfeat} & 19/\textbf{0.19} & \textbf{12}/\textbf{0.25} & 3.9/0.17& 7.2/0.24& \textbf{11}/\textbf{0.21} \\
  & ELoFTR~\cite{wang2024efficient} & 19/0.20 & 13/\textbf{0.25} & 3.7/\textbf{0.15}& 7.0/0.23 & \textbf{11}/\textbf{0.21} \\
  & MatchAny~\cite{li2024matching} & 19/0.20 & 13/0.26 & \textbf{3.6}/0.17& 6.6/\textbf{0.20} & \textbf{11}/\textbf{0.21} \\
  & MASt3r~\cite{leroy2024grounding} & \textbf{18}/\textbf{0.19} & 14/0.30 & 4.5/0.16& \textbf{6.5}/0.21& \textbf{11}/0.22 \\

\bottomrule 
\end{tabular}
}
\end{threeparttable}

\label{tab:supp_matcher}
\end{table*}

\begin{table*}[ht!]
\scriptsize
\centering
\caption{\textbf{Module Ablation on Cambridge Landmark.} We report the median translation error (cm) and rotation error ($^\circ$) on Cambridge Landmark dataset~\cite{kendall2015posenet}. We utilize DFNet~\cite{chen2022dfnet} and ACE~\cite{brachmann2023accelerated} as a pose prior for module ablation. MCR and UPnP denote Monte Carlo refinement and Uncertainty sampling-based PnP optimization.
}
\setlength{\tabcolsep}{4pt} %
\begin{threeparttable}
\resizebox{0.7\textwidth}{!}
{
\begin{tabular}{cccccccc}
\toprule
Prior & MCR & UPnP & Kings  & Hospital  & Shop & Church  &Avg. $\downarrow$ [$\text{cm}/^\circ$]   \\\midrule
\multirow{4}{*}{DFNet} &  &  & 21/0.25 & 29/0.54 & 9.4/0.35 & 14/0.38& 19/0.38 \\
 &  &  \checkmark & 20/0.25 & 25/0.58 & 9.2/0.34 & 14/0.39& 17/0.39\\
 & \checkmark &  & \textbf{19}/0.20 & 17/\textbf{0.30} & 4.4/0.18& 5.8/\textbf{0.18}& 12/\textbf{0.22} \\
 & \checkmark & \checkmark & \textbf{19}/\textbf{0.19} & \textbf{15}/\textbf{0.30} & \textbf{4.3}/\textbf{0.17}& \textbf{5.7}/\textbf{0.18}& \textbf{11}/\textbf{0.22} \\
 \midrule
\multirow{4}{*}{ACE} &  &  & 21/0.25 & 19 /0.36 & 4.7/0.20& 9.5/0.29& 14/0.28 \\
  &  & \checkmark & 19/0.21 & 18 /0.33 & 4.6/0.20& 8.3/0.26& 12/0.25 \\
  & \checkmark & & \textbf{18}/0.20 & 15 /0.31 & 4.3/0.18& 6.8/\textbf{0.2}1& \textbf{11}/0.23 \\
  & \checkmark & \checkmark & \textbf{18}/\textbf{0.19} & \textbf{14} /\textbf{0.29} &\textbf{ 4.3}/\textbf{0.17}& \textbf{6.3}/\textbf{0.21}& \textbf{11}/\textbf{0.22} \\

\bottomrule 
\end{tabular}
}
\end{threeparttable}

\label{tab:supp_module_abl_camb}
\end{table*}

\begin{table*}[ht!]
\scriptsize
\centering
\caption{\textbf{Module Ablation on 7scenes~\cite{shotton2013scene}.} We report the median translation error (cm) and rotation error ($^\circ$) on 7-scenes dataset. We utilize DFNet~\cite{chen2022dfnet} as a pose prior for module ablation. MCR and UPnP denote Monte Carlo refinement and Uncertainty sampling-based PnP optimization.
}
\setlength{\tabcolsep}{4pt} %
\begin{threeparttable}
\resizebox{\textwidth}{!}
{
\begin{tabular}{ccccccccccc}
\toprule
 Prior & MCR & UPnP & Chess  & Fire  & Heads & Office & Pumpkin  & Redkitchen & Stairs  & Avg. $\downarrow$ [$\text{cm}/^\circ$] \\\midrule
 \multirow{4}{*}{DFNet} &  &  & 0.63/0.18 & 0.93/0.33 & 0.56/0.36& 1.27/0.33 & 1.18/0.28 & 0.93/0.24 & 2.18/0.59 & 1.10/0.33 \\
  &  & \checkmark  & 0.53/0.16 & 0.81/0.31 & 0.58/0.38 & 1.10/0.29 & 0.99/0.23 & 0.85/0.23& 2.40/0.63& 1.04/0.32 \\
  & \checkmark &  & 0.40/0.13& 0.55/0.21 & 0.38/\textbf{0.25}& 0.83/0.23 & 0.89/0.21 & 0.63/\textbf{0.15} & \textbf{1.18}/\textbf{0.35} & 0.69/0.22 \\
  & \checkmark & \checkmark & \textbf{0.36}/\textbf{0.12}& \textbf{0.49}/\textbf{0.20} & \textbf{0.36}/\textbf{0.25}& \textbf{0.78}/\textbf{0.22} & \textbf{0.81}/\textbf{0.18} & \textbf{0.59}/\textbf{0.15} & 1.19/\textbf{0.35}& \textbf{0.65}/\textbf{0.21} \\

\bottomrule 
\end{tabular}
}
\end{threeparttable}

\label{tab:supp_module_abl_7scenes}
\end{table*}
\vspace{-0.3cm}
\section{Module Ablation}

\cref{tab:supp_module_abl_camb,tab:supp_module_abl_7scenes} report how each component of our pipeline, Monte Carlo Refinement (MCR) and Uncertainty-based PnP (UPnP), contributes to the final localization accuracy. We evaluate both DFNet~\cite{chen2022dfnet} and ACE~\cite{brachmann2023accelerated} as pose priors to analyze the generality of each module across different initialization qualities.
For both priors, enabling only UPnP yields modest improvements by suppressing unreliable correspondences through depth-uncertainty guidance. Incorporating only MCR produces a larger gain, reflecting the benefit of exploring multiple pose hypotheses rather than relying on a single deterministic prior. 

As shown in \cref{tab:supp_module_abl_camb}, applying our uncertainty-based PnP alone already improves the median translation error by roughly 11\% compared to the standard 3DGS-based pose refinement. 
When combined with the modified Monte Carlo refinement, the performance further improves with two iterations, the MCL structure achieves a substantial reduction in both translation and rotation errors. When the two modules are combined, we observe the best performance across all scenes. Under DFNet pose prior, the combined system reduces the average error to 11cm and 0.22$^\circ$, which is the same performace utilizing a better pose prior (ACE). This consistent convergence across different priors demonstrates that each module addresses a distinct source of uncertainty. MCR mitigates pose prior bias, whereas UPnP handles geometric unreliability during 2D–3D lifting.

Overall, the ablation confirms that both components contribute complementary strengths, and their combination forms a robust and reliable pose refinement pipeline independent of the choice of pose estimator.

\begin{figure*}[!th]
  \centering
  \includegraphics[width=\textwidth]{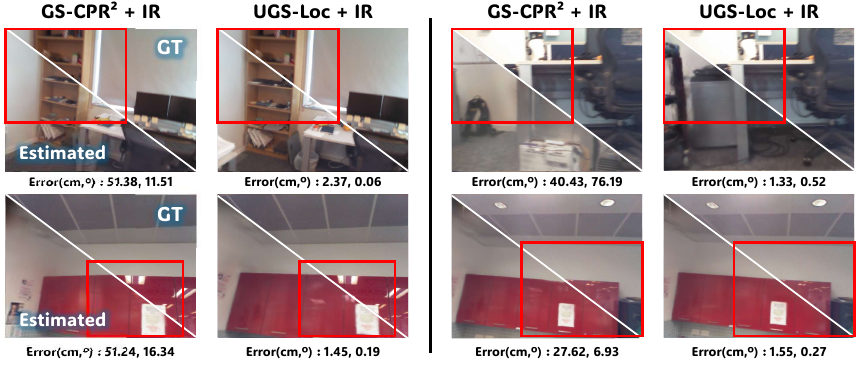}
\caption{\textbf{Visualization of Localization Quality on the 7-Scenes Dataset.}
Each pair of columns compares the ground-truth view (top-right) against the view rendered from 
(i) the iteratively modified GS-CPR refinement with 2 iterations (denoted as GS-CPR$^2$)~\cite{liu2024gs} and 
(ii) our UGS-Loc refinement (bottom-left). IR denotes that the pose prior is initialized with Image Retrieval (IR)~\cite{arandjelovic2016netvlad}. 
}
\label{fig:supp_IR_7scenes}
\end{figure*}

\section{Inference Cost}

We do not claim that UGS-Loc offers faster inference than deterministic 3DGS-based pose refinement. Because our framework integrates the conventional 3DGS refinement pipeline with an MCL-inspired multi-hypothesis strategy, additional computation is naturally introduced. However, rather than executing iterative refinement serially as in traditional 3DGS pipelines, UGS-Loc leverages \textit{multiprocessing} to evaluate multiple pose hypotheses in parallel. This design helps mitigate much of the computational overhead and yields practical efficiency in multi-particle settings. Another important factor influencing inference speed is the choice of the 2D matching module. As discussed in \cref{sec:supp_match}, UGS-Loc maintains strong localization performance even when paired with lightweight matchers such as XFeat or SP+LG, providing flexibility for balancing speed and accuracy beyond the MASt3r-based configuration.

\paragraph{Comparison with GS-CPR.}
Due to differences in hardware, a direct inference time comparison with GS-CPR~\cite{liu2024gs} requires normalization. According to the original GS-CPR paper, the runtime of each component is as follows: 3.7ms for a single rendering, 71ms for a MASt3r~\cite{leroy2024grounding} forward pass, and 94 ms for the PnP optimization with MASt3r matching. However, under our environment, using the same codebase, rendering, MASt3r inference, and PnP optimization take 12.4ms, 189ms, and 182ms, respectively, showing a consistent slowdown across all processes. This discrepancy arises from hardware differences rather than algorithmic overhead. 

\begin{figure*}[!th]
  \centering
  \includegraphics[width=\textwidth]{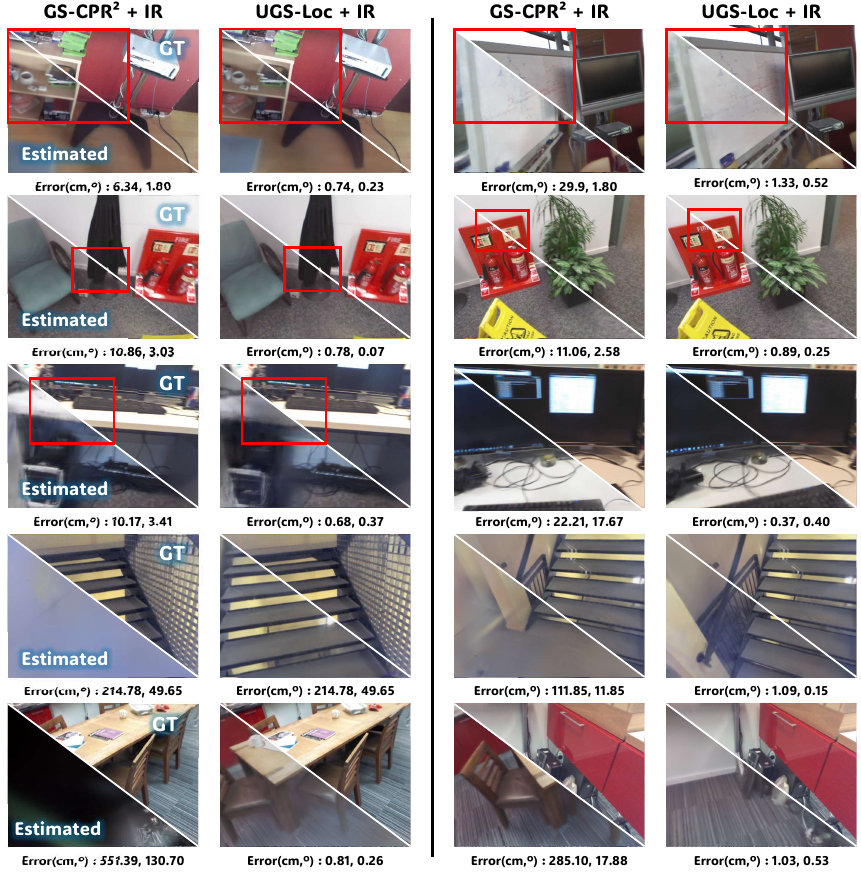}
\caption{\textbf{Visualization of Localization Quality on the 7-Scenes Dataset.}
Each pair of columns compares the ground-truth view (top-right) against the view rendered from 
(i) the iteratively modified GS-CPR refinement with 2 iterations (denoted as GS-CPR$^2$)~\cite{liu2024gs} and 
(ii) our UGS-Loc refinement (bottom-left). IR denotes that the pose prior is initialized with Image Retrieval (IR)~\cite{arandjelovic2016netvlad}. 
}
\label{fig:supp_IR_7scenes}
\end{figure*}

\section{Additional Implementation Detail}

This section addresses the hyperparameters and implementation details used in our pose refinement pipeline. UGS-Loc performs PnP-based refinement over two iterations, maintaining eight particles per iteration. For efficiency, we use a 256 resolution in the first iteration and a 512 resolution in the second. For PnP optimization, the reprojection error threshold is set to 1.0 px for 7Scenes and 2.5 px or Cambridge. The maximum number of iterations for PnP-RANSAC is fixed to 1000, consistent with GS-CPR~\cite{liu2024gs}. Our uncertainty-guided sampling–based PnP terminates early once the confidence threshold reaches 0.99. To prevent extreme values in correspondence uncertainty values $u_i$, we apply percentile clipping using the 5\%–95\% range. During the final iteration of Monte Carlo refinement, we re-render all particles from their refined poses and compute SSIM scores~\cite{wang2004image} to obtain the particle weights. For 7Scenes, the final pose is obtained via a weighted average, whereas for Cambridge, we select the best single particle to avoid performance degradation caused by noisy samples.

For scene representation, we adopt Scaffold-GS~\cite{lu2024scaffold} and apply object and sky masks obtained from an off-the-shelf model~\cite{cheng2022masked} to prevent overly noisy Gaussian reconstructions. Unlike several prior localization works~\cite{huang2025sparse,li2024unleashing}, our 7Scenes experiments train the Gaussian Splatting model using RGB images only, without requiring depth supervision or per-scene fine-tuning with deep features beyond GS training.